\documentclass[11pt,a4paper]{article}
\usepackage[hyperref]{acl2019}
\usepackage{times}
\usepackage{latexsym}
\usepackage{url}
\usepackage{CJKutf8}
\usepackage{booktabs}
\usepackage{makecell}
\usepackage{multirow}
\usepackage{amssymb}
\usepackage{amsmath}
\usepackage{caption}
\usepackage{graphicx, subfig}
\usepackage{color}
\usepackage{hyperref}

\aclfinalcopy 

 \title{Modeling Semantic Compositionality with Sememe Knowledge}

\author{Fanchao Qi$^{1}\thanks{\ \  Indicates equal contribution}$\hspace{0.2em}, 
Junjie Huang$^{2*}\thanks{\ \ Work done during internship at Tsinghua University}$\hspace{0.2em}, 
Chenghao Yang$^{3\dag}$, 
Zhiyuan Liu$^{1}$, \\
{\bf Xiao Chen$^{4}$,
Qun Liu$^{4}$,
Maosong Sun$^{1}$\thanks{\ \  Corresponding author}}\hspace{0.5em}\\
$^{1}$Department of Computer Science and Technology, Tsinghua University \\
Institute for Artificial Intelligence, Tsinghua University \\
State Key Lab on Intelligent Technology and Systems, Tsinghua University\\
$^{2}$School of ASEE, Beihang University \quad
$^{3}$Software College, Beihang Unviersity\\
$^{4}$Huawei Noah's Ark Lab\\
{\tt qfc17@mails.tsinghua.edu.cn, \{hjj1997,alanyang\}@buaa.edu.cn}\\
{\tt \{liuzy,sms\}@tsinghua.edu.cn, \{chen.xiao2,qun.liu\}@huawei.com}
}

\date{}

\begin{document}
\begin{CJK}{UTF8}{gkai}
\maketitle
\begin{abstract}

Semantic compositionality (SC) refers to the phenomenon that the meaning of a complex linguistic unit can be composed of the meanings of its constituents. 
Most related works focus on using complicated compositionality functions to model SC while few works consider external knowledge in models.
In this paper, we verify the effectiveness of sememes, the minimum semantic units of human languages, in modeling SC by a confirmatory experiment. 
Furthermore, we make the first attempt to incorporate sememe knowledge into SC models, and employ the sememe-incorporated models in learning representations of multiword expressions, a typical task of SC. 
In experiments, we implement our models by incorporating knowledge from a famous sememe knowledge base HowNet and perform both intrinsic and extrinsic evaluations. 
Experimental results show that our models achieve significant performance boost as compared to the baseline methods without considering sememe knowledge. 
We further conduct quantitative analysis and case studies to demonstrate the effectiveness of applying sememe knowledge in modeling SC.
All the code and data of this paper can be obtained on \url{https://github.com/thunlp/Sememe-SC}.
 
\end{abstract}

\section{Introduction}

Semantic compositionality (SC) is defined as the linguistic phenomenon that the meaning of a syntactically complex unit is a function of meanings of the complex unit's constituents and their combination rule \citep{Pelletier1994}. 
Some linguists regard SC as the fundamental truth of semantics \citep{Pelletier2016}. 
In the field of NLP, SC has proved effective in many tasks including language modeling \citep{Mitchell2009}, sentiment analysis \citep{maas2011learning,Socher2013}, syntactic parsing \citep{socher2013parsing}, etc.

Most literature on SC pays attention to using vector-based distributional models of semantics to learn representations of multiword expressions (MWEs), i.e., embeddings of phrases or compounds. \citet{Mitchell2008} conduct a pioneering work in which they introduce a general framework to formulate this task:
\begin{equation}
\label{eq:composition}
	\textbf{p}=f(\textbf{w}_1,\textbf{w}_2,R,K)\footnote{This formula only applies to two-word MWEs but can be easily extended to longer MWEs. In fact, we also focus on modeling SC for two-word MWEs in this paper because they are the most common.},
\end{equation} 
where $f$ is the compositionality function, \textbf{p} denotes the embedding of an MWE, 
$\textbf{w}_1$ and $\textbf{w}_2$ represent the embeddings of the MWE's two constituents, $R$ stands for the combination rule and $K$ refers to the additional knowledge which is needed to construct the semantics of the MWE. 


Among the proposed approaches for this task, most of them ignore $R$ and $K$, centering on reforming compositionality function $f$ \citep{baroni2010nouns,grefenstette2011experimental,Socher2012,Socher2013}. 
Some try to integrate combination rule $R$ into SC models \citep{Blacoe2012,Zhao2015,Weir2016,Kober2016}. 
Few works consider external knowledge $K$.
\citet{Zhu2016} try to incorporate task-specific knowledge into an LSTM model for sentence-level SC.
As far as we know, however, no previous work attempts to use general knowledge in modeling SC.

\begin{table*}[t!]
\small
  \centering
  \resizebox{1.0\textwidth}{!}{
    \begin{tabular}{cclcccc}
    \toprule
    \multicolumn{1}{c}{\multirow{2}[4]{*}{\makecell{SCD}}} & \multirow{2}[4]{*}{Our Computation Formulae} & \multicolumn{5}{c}{Examples} \\
\cmidrule{3-7}          &       & \multicolumn{1}{c}{MWEs and Constituents} & \multicolumn{4}{c}{Sememes} \\

    \midrule
    \multirow{3}[2]{*}{3} & \multirow{3}[2]{*}{\large{$S_{\boldsymbol{p}} = S_{\boldsymbol{w}_{1}} \cup S_{\boldsymbol{w}_{2}}$}} & \textcolor{red}{农民}\textcolor{blue}{起义}(peasant uprising) & \multicolumn{4}{l}{事情$|$fact,职位$|$occupation,政$|$politics,暴动$|$uprise,人$|$human,农$|$agricultural} \\
          &       & \textcolor{red}{农民}\qquad (peasant) & \multicolumn{4}{l}{\textbf{职位$|$occupation},\textbf{人$|$human},\textbf{农$|$agricultural}} \\
          &       & \qquad \textcolor{blue}{起义}(uprising) & \multicolumn{4}{l}{\textbf{暴动$|$uprise},\textbf{事情$|$fact},\textbf{政$|$politics}} \\
    \midrule
    \multirow{3}[2]{*}{2} & \multirow{3}[2]{*}{\large{$S_{\boldsymbol{p}} \subsetneq (S_{\boldsymbol{w}_{1}} \cup S_{\boldsymbol{w}_{2}})$}} & \textcolor{red}{几何}\textcolor{blue}{图形}(geometric figure) & \multicolumn{4}{l}{数学$|$math,图像$|$image} \\
          &       & \textcolor{red}{几何}\qquad (geometry; how much)  & \multicolumn{4}{l}{\textbf{数学$|$math},知识$|$knowledge,疑问$|$question,功能词$|$funcword} \\
          &       & \qquad \textcolor{blue}{图形}(figure) & \multicolumn{4}{l}{\textbf{图像$|$image}} \\
    \midrule
    \multirow{3}[2]{*}{1} & \multirow{3}[2]{*}{\large{$\makecell{S_{\boldsymbol{p}} \cap (S_{\boldsymbol{w}_{1}} \cup S_{\boldsymbol{w}_{2}}) \neq \emptyset \\  \land\ S_{\boldsymbol{p}} \not\subset (S_{\boldsymbol{w}_{1}} \cup S_{\boldsymbol{w}_{2}})}$}} & \textcolor{red}{应}\textcolor{blue}{考}(engage a test) & \multicolumn{4}{l}{考试$|$exam,从事$|$engage} \\
          &       & \textcolor{red}{应}\quad (deal with; echo; agree) & \multicolumn{4}{l}{处理$|$handle,回应$|$respond,同意$|$agree,遵循$|$obey,功能词$|$funcword,姓$|$surname} \\
          &       & \quad \textcolor{blue}{考}(quiz; check) & \multicolumn{4}{l}{\textbf{考试$|$exam},查$|$check} \\
    \midrule
    \multirow{3}[2]{*}{0} & \multirow{3}[2]{*}{\large{$S_{\boldsymbol{p}} \cap (S_{\boldsymbol{w}_{1}} \cup S_{\boldsymbol{w}_{2}}) = \emptyset$}} & \textcolor{red}{画}\textcolor{blue}{句号}(end) & \multicolumn{4}{l}{完毕$|$finish} \\
          &       & \textcolor{red}{画}\qquad (draw) & \multicolumn{4}{l}{画$|$draw,部件$|$part,图像$|$image, 文字$|$character,表示$|$express} \\
          &       & \quad \textcolor{blue}{句号}(period) & \multicolumn{4}{l}{符号$|$symbol,语文$|$text} \\
    \bottomrule
    \end{tabular}
}
\caption{Sememe-based semantic compositionality degree computation formulae and examples. Bold sememes of constituents are shared with the constituents' corresponding MWE.}
  \label{tab:compositionlity_score}%
\end{table*}%

In fact, there exists general linguistic knowledge which can be used in modeling SC, e.g., sememes. Sememes are defined as the minimum semantic units of human languages \citep{bloomfield1926set}. 
It is believed that the meanings of all the words can be composed of a limited set of sememes, which is similar to the idea of \textit{semantic primes} \citep{wierzbicka1996semantics}. HowNet \citep{dong2003hownet} is a widely acknowledged sememe knowledge base (KB), which defines about 2,000 sememes and uses them to annotate over 100,000 Chinese words together with their English translations. Sememes and HowNet have been successfully utilized in a variety of NLP tasks including sentiment analysis \citep{dang2010method}, word representation learning \citep{niu2017improved}, language modeling \citep{gu2018language}, etc. 

In this paper, we argue that sememes are beneficial to modeling SC\footnote{Since HowNet mainly annotates Chinese words with sememes, we experiment on Chinese MWEs in this paper. But our methods and findings are also applicable to other languages.}. 
To verify this, we first design a simple SC degree (SCD) measurement experiment and find that the SCDs of MWEs computed by simple sememe-based formulae are highly correlated with human judgment. 
This result shows that sememes can finely depict meanings of MWEs and their constituents, and capture the semantic relations between the two sides. 
Therefore, we believe that sememes are appropriate for modeling SC and can improve the performance of SC-related tasks like MWE representation learning.

We propose two sememe-incorporated SC models for learning embeddings of MWEs, namely Semantic Compositionality with Aggregated Sememe (SCAS) model and Semantic Compositionality with Mutual Sememe Attention (SCMSA) model. 
When learning the embedding of an MWE, SCAS model concatenates the embeddings of the MWE's constituents and their sememes, while SCMSA model considers the mutual attention between a constituent's sememes and the other constituent. 
We also integrate the combination rule, i.e., $R$ in Eq. \eqref{eq:composition}, into the two models.
We evaluate our models on the task of MWE similarity computation, finding our models obtain significant performance improvement as compared to baseline methods.
Furthermore, we propose to evaluate SC models on a downstream task sememe prediction, and our models also exhibit favorable outcomes. 


\section{Measuring SC Degree with Sememes}
\label{sec:sec2}
In this section, we conduct a confirmatory SCD measurement experiment to present evidence that sememes are appropriate for modeling SC.



\subsection{Sememe-based SCD Computation Formulae}
Although SC widely exists in MWEs, not every MWE is fully semantically compositional. In fact, different MWEs show different degrees of SC. 
We believe that sememes can be used to measure SCD conveniently. 
To this end, based on the assumption that all the sememes of a word accurately depict the word's meaning, we intuitively design a set of SCD computation formulae, which we believe are consistent with the principle of SCD. 

The formulae are illustrated in Table \ref{tab:compositionlity_score}. 
We define four SCDs denoted by number 3, 2, 1 and 0, where larger numbers mean higher SCDs. 
$S_{\boldsymbol{p}}$, $S_{\boldsymbol{w}_{1}}$ and $S_{\boldsymbol{w}_{2}}$ represent the sememe sets of an MWE, its first and second constituent respectively.

Next, we give a brief explanation for these SCD computation formulae: 
(1) For SCD 3, 
the sememe set of an MWE is identical to the union of the two constituents' sememe sets, which means the meaning of the MWE is exactly the same as the combination of the constituents' meanings. Therefore, the MWE is fully semantically compositional and should have the highest SCD. 
(2) For SCD 0, an MWE has totally different sememes from its constituents, which means the MWE's meaning cannot be derived from its constituents' meanings. Hence the MWE is completely non-compositional, and its SCD should be the lowest.
(3) As for SCD 2, the sememe set of an MWE is a proper subset of the union of its constituents' sememe sets, which means the meanings of the constituents cover the MWE's meaning but cannot precisely infer the MWE's meaning. 
(4) Finally, for SCD 1, 
an MWE shares some sememes with its constituents, but both the MWE itself and its constituents have some unique sememes. 

In Table \ref{tab:compositionlity_score}, we also show an example for each SCD, including a Chinese MWE, its two constituents and their sememes\footnote{In Chinese, most MWEs are words consisting of more than two characters which are actually single-morpheme words.}. 

\subsection{Evaluating SCD Computation Formulae}
To evaluate our sememe-based SCD computation formulae, we construct a human-annotated SCD dataset. 
We ask several native speakers to label SCDs for 500 Chinese MWEs, where there are also four degrees to choose. 
Before labeling an MWE, they are shown the dictionary definitions of both the MWE and its constituents. 
Each MWE is labeled by 3 annotators, and the average of the 3 SCDs given by them is the MWE's final SCD.
Eventually, we obtain a dataset containing 500 Chinese MWEs together with their human-annotated SCDs. 

Then we evaluate the correlativity between SCDs of the MWEs in the dataset computed by sememe-based rules and those given by humans. 
We find Pearson's correlation coefficient is up to \textbf{0.75}, and Spearman's rank correlation coefficient is 0.74. 
These results manifest remarkable capability of sememes to compute SCDs of MWEs and provide proof that sememes of a word can finely represent the word's meaning. 
Accordingly, we believe that this characteristic of sememes can also be exploited in modeling SC. 


 


 

\section{Sememe-incorporated SC Models}
In this section, we first introduce our two basic sememe-incorporated SC models in detail, namely Semantic Compositionality with Aggregated Sememe (SCAS) and Semantic Compositionality with Mutual Sememe Attention (SCMSA). 
SCAS model simply concatenates the embeddings of the MWE’s constituents and their sememes, while SCMSA model takes account of the mutual attention between a constituent’s sememes and the other constituent. 
Then we describe how to integrate combination rules into the two basic models. 
Finally, we present the training strategies and losses for two different tasks.
 
 

\subsection{Incorporating Sememes Only} 

Following the notations in Eq. \eqref{eq:composition}, for an MWE $p = \{w_1, w_2\}$, its embedding can be represented as:
\begin{equation}
\mathbf{p} = f(\mathbf{w}_{1}, \mathbf{w}_{2}, K),
\label{phrase_func_word_knowledge}
\end{equation}
where $\mathbf{p},\mathbf{w}_1,\mathbf{w}_2 \in \mathbb{R}^{d}$ and $d$ is the dimension of embeddings. 
$K$ denotes the sememe knowledge here, and we assume that we only know the sememes of $w_1$ and $w_2$, considering that MWEs are normally not in the sememe KBs. 
We use $S$ to indicate the set of all the sememes and $S_w=\{s_1,...,s_{|S_w|}\}\subset S$ to signify the sememe set of $w$, where $|\cdot|$ represents the cardinality of a set. 
In addition, $\mathbf{s}\in \mathbb{R}^{d}$ denotes the embedding of sememe $s$.




\begin{figure}[!t]
\centering
\includegraphics[scale=0.45]{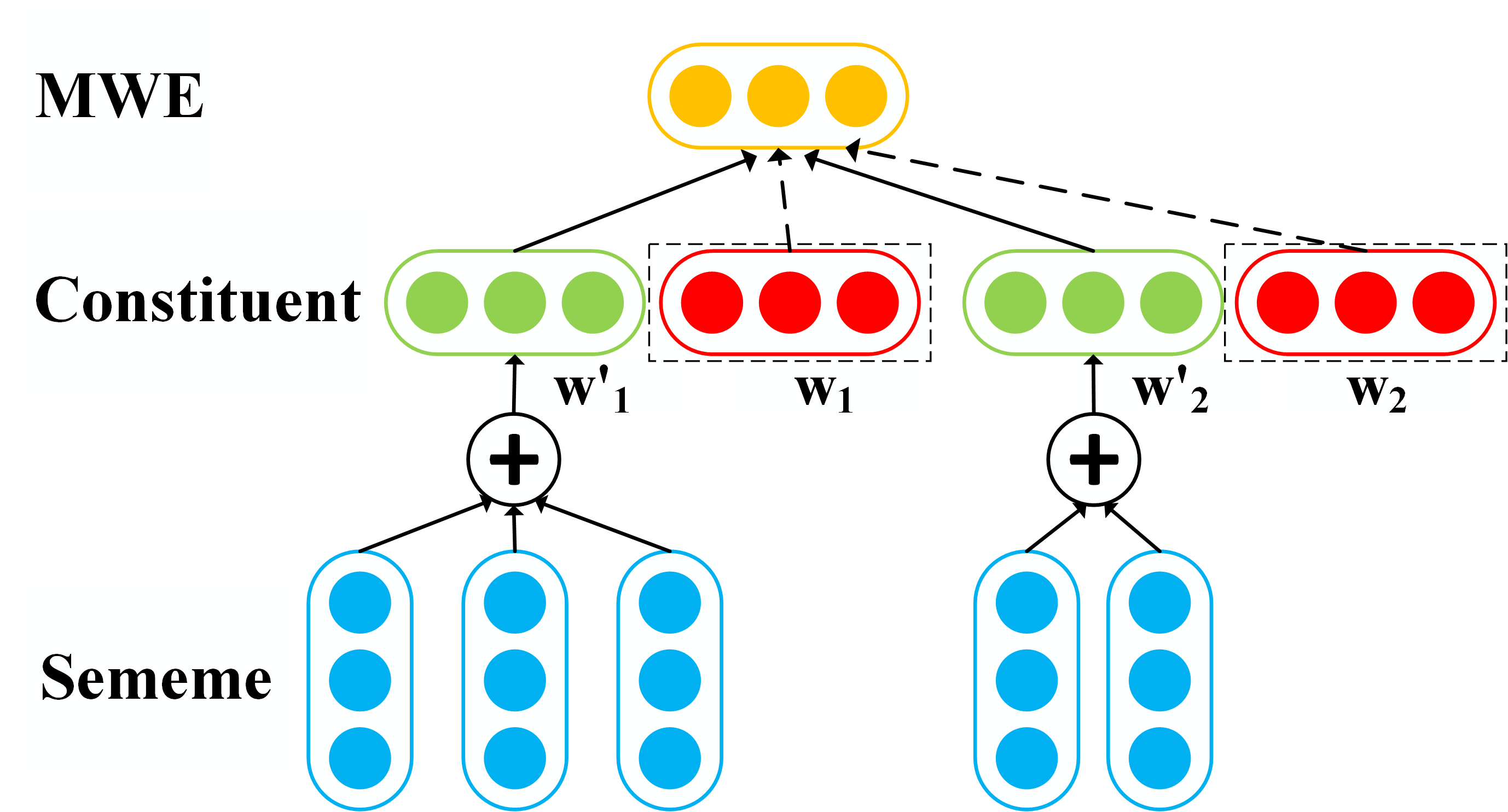}
\caption{Semantic Compositionality with Aggregated Sememe (SCAS) model.}
\label{fig:model1}
\end{figure}

\subsubsection*{SCAS Model}
The first model we propose is SCAS model, which is illustrated in Figure \ref{fig:model1}. 
The idea of SCAS model is straightforward, i.e., simply
concatenating word embedding of a constituent and the aggregation of its sememes' embeddings.
Formally, we have:
\begin{equation}
\mathbf{w}_{1}^{'} =  \sum_{s_i \in S_{w_{1}}} \mathbf{s_i},  
\quad
\mathbf{w}_{2}^{'} = \sum_{s_j \in S_{w_{2}}} \mathbf{s_j},
\label{sememe_word_left}
\end{equation}
where $\mathbf{w}_{1}^{'}$ and $\mathbf{w}_{2}^{'}$ represent the aggregated sememe embeddings of $w_1$ and $w_2$ respectively.
Then $\mathbf{p}$ can be obtained by:
\begin{equation}
\mathbf{p} = \tanh(\mathbf{W}_c \lbrack \mathbf{w}_{1}+\mathbf{w}_{2}  \text{;} \mathbf{w}_{1}^{'}+\mathbf{w}_{2}^{'} \rbrack + \mathbf{b}_c),
\label{eq:compute_p}
\end{equation}
where $\mathbf{W}_c \in \mathbb{R}^{d \times 2d}$ is the composition matrix and $\mathbf{b}_c \in \mathbb{R}^{d}$ is a bias vector.

\subsubsection*{SCMSA Model}

\begin{figure}[!t]
\centering
\includegraphics[scale=0.45]{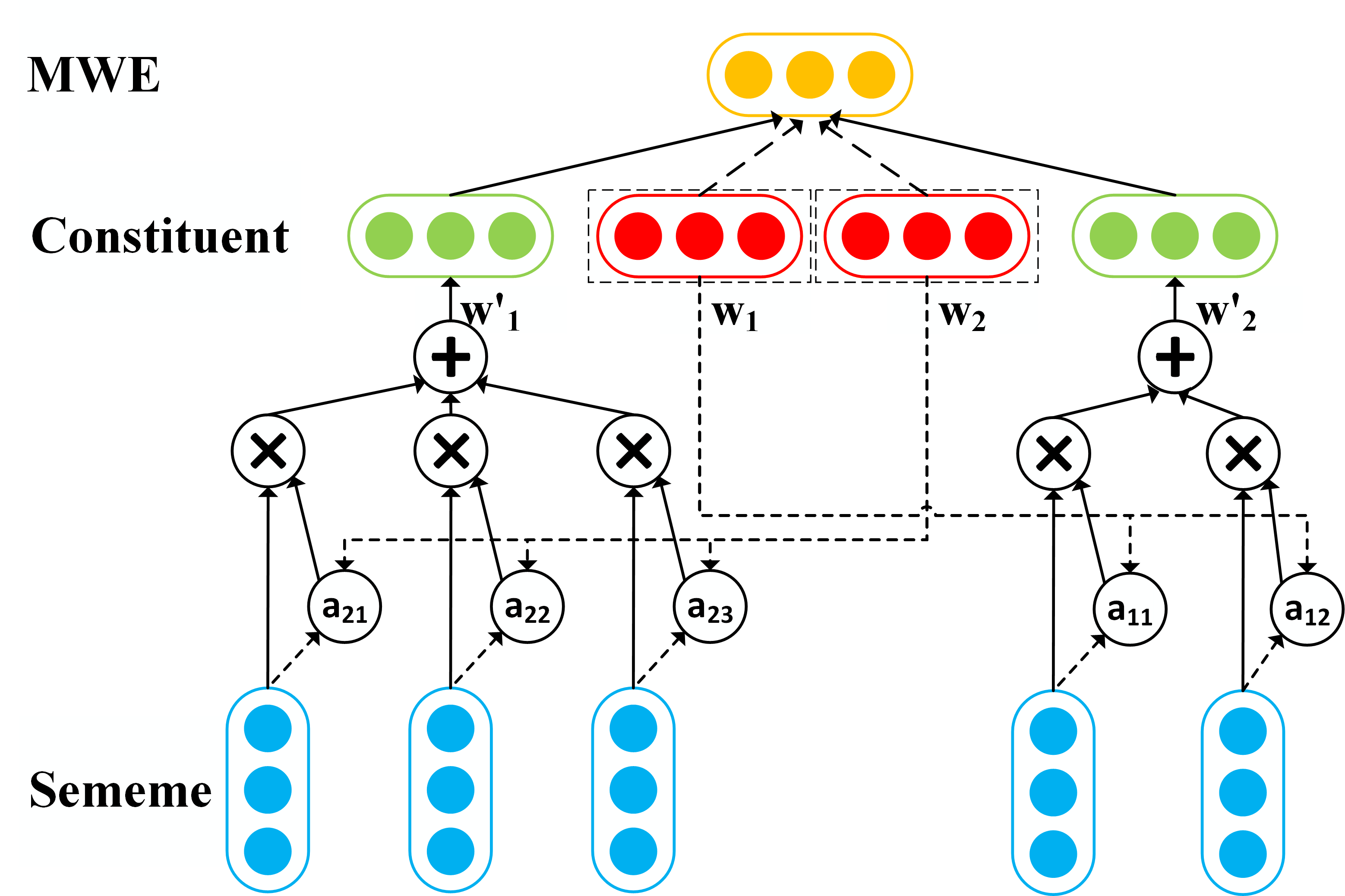}
\caption{Semantic Compositionality with Mutual Sememe Attention (SCMSA) model.}
\label{fig:model2}
\end{figure}

The SCAS model simply uses the sum of all the sememes' embeddings of a constituent as the external information. 
However, a constituent's meaning may vary with the other constituent, and accordingly, the sememes of a constituent should have different weights when the constituent is combined with different constituents (we show an example in later case study). 
Correspondingly, we propose SCMSA model (Figure \ref{fig:model2}), which adopts the mutual attention mechanism to dynamically endow sememes with weights.



Formally, we have:
\begin{equation}
\begin{aligned}
    \mathbf{e}_{1} &=  \tanh(\mathbf{W}_a \mathbf{w}_{1} + \mathbf{b}_a), \\
    a_{2,i} &= \frac{\exp{(\mathbf{s_i} \cdot \mathbf{e}_{1})}}{\sum_{s_j \in S_{w_{2}}} \exp{(\mathbf{s_j} \cdot \mathbf{e}_{1})}},\\
    \mathbf{w}_{2}^{'} &= \sum_{s_i \in S_{w_{2}}} a_{2,i} \mathbf{s_i},
\end{aligned}
\label{eq:w1_on_w2sememe}
\end{equation}
where $\mathbf{W}_a \in \mathbb{R}^{d \times d}$ is the weight matrix and $\mathbf{b}_a \in \mathbb{R}^{d}$ is a bias vector.

Similarly, we can calculate $\mathbf{w}'_1$. 
Then we still use Eq. \eqref{eq:compute_p} to obtain $\mathbf{p}$.

\subsection{Integrating Combination Rules}
In this section, we further integrate combination rules into our sememe-incorporated SC models. In other words, 
\begin{equation}
\mathbf{p} = f(\mathbf{w}_{1}, \mathbf{w}_{2},  K, R).
\label{phrase_func_word_knowledge}
\end{equation}

We can use totally different composition matrices for MWEs with different combination rules:
\begin{equation}
    \mathbf{W}_c = \mathbf{W}_c^r, \quad r\in R_s
\end{equation}
where $\mathbf{W}_c^r \in \mathbb{R}^{d \times 2d}$ and $R_s$ refers to combination rule set containing syntax rules of MWEs, e.g., adjective-noun and noun-noun. 

However, there are many different combination rules and  some rules have sparse instances which are not enough to train the corresponding composition matrices with $d \times 2d$ parameters. In addition, we believe that the composition matrix should contain common compositionality information except the combination rule-specific compositionality information.
Hence we let composition matrix $\mathbf{W}_c$ be the sum of a low-rank matrix containing combination rule information and a matrix containing common compositionality information:
\begin{equation}
\label{eq:rule}
    \mathbf{W}_c = \mathbf{U}^r \mathbf{V}^r + \mathbf{W}^c_{c},
\end{equation}
where $\mathbf{U}^r \in \mathbb{R}^{d \times h_r}$,  $\mathbf{V}^r \in \mathbb{R}^{h_r \times 2d}$, $h_r\in \mathbb{N}_+$ is a hyper-parameter and may vary with the combination rule, 
and $\textbf{W}^c_{c} \in \mathbb{R}^{d \times 2d}$.

\subsection{Training}
We use the MWE embeddings obtained by above-mentioned SC models in downstream tasks. 
For different tasks, we adopt different training strategies and loss functions.


\subsubsection*{Training for MWE Similarity Computation}
For the task of MWE similarity computation, we use the squared Euclidean distance loss following \citet{Luong2013}. 
For an MWE $p$, its training loss is: 
\begin{equation}
    L_p = \left\|\mathbf{p}^c - \mathbf{p}^r\right\|_2^2,
\end{equation}
where $\mathbf{p}^c \in \mathbb{R}^d$ is the embedding of $p$ obtained by our SC models
, i.e.,  previous $\mathbf{p}$, 
and $\mathbf{p}^r \in \mathbb{R}^d$ is the corresponding reference embedding, which might be obtained by regarding the MWE as a whole and applying word representation learning methods. 

And the overall loss function is as follows:
\begin{equation}
    L = \sum_{p\in P_t}L_p+ \frac{\lambda}{2}\sum_{\theta\in\Theta}\left\|\theta\right\|_2^2,
\end{equation}
where $P_t$ is the training set, $\Theta$ refers to the parameter set including $\mathbf{W}_c$ and $\mathbf{W}_a$, and $\lambda$ is the regularization parameter.



\subsubsection*{Training for MWE Sememe Prediction}
\label{sec:MWE_sp}
Sememe prediction is a well-defined task \citep{xie2017lexical,jin2018incorporating,qi2018cross}, aimed at selecting appropriate sememes for unannotated words or phrases from the set of all the sememes.
Existing works model sememe prediction as a multi-label classification problem, where sememes are regarded as the labels of words and phrases. 


For doing MWE sememe prediction, we employ a single-layer perceptron as the classifier:
\begin{equation}
    \hat{\mathbf{y}}_p = \sigma(\mathbf{W}_s \cdot \mathbf{p}),
\end{equation}
where $\hat{\mathbf{y}}_p \in\mathbb{R}^{|S|}$, $\mathbf{W}_s\in \mathbb{R}^{|S|\times d}$ and $\sigma$ is the sigmoid function. 
$[\hat{\mathbf{y}}_p]_i$, the $i$-th element of $\hat{\mathbf{y}}_p$, denotes the predicted score of $i$-th sememe, where the higher the score is, the more probable the sememe is selected. 
And $\mathbf{W}_s = [\mathbf{s}_1,\cdots,\mathbf{s}_{|S|}]^{\top}$ is made up of the embeddings of all the sememes.

As for the training loss of the classifier, considering the distribution of sememes over words is quite imbalanced, we adopt the weighted cross-entropy loss: 
\begin{equation}
\label{eq:loss2}
\begin{aligned}
    L = & \sum_{p\in P_t} \sum_{i=1}^{|S|} \big( k \times [\mathbf{y}_p]_i\log[\hat{\mathbf{y}}_p]_i  \\ 
    & + (1-[\mathbf{y}_p]_i)\log(1-[\hat{\mathbf{y}}_p]_i) \big) ,
\end{aligned}
\end{equation}
where $[\mathbf{y}_p]_i \in \{0, 1\}$ is the $i$-th element of $\mathbf{y}_p$, which is the true sememe label of $p$, and $k$ stands for the weight parameter.



\section{Experiments}
We evaluate our sememe-incorporated SC models on two tasks including MWE similarity computation and MWE sememe prediction. For the latter, we also conduct further quantitative analysis and case study.


\subsection{Dataset}


We choose HowNet as the source of sememe knowledge. In HowNet, there are 118,346 Chinese words annotated with 2,138 sememes in total. Following previous work \citep{xie2017lexical,jin2018incorporating}, we filter out the low-frequency sememes, which are considered unimportant. The final number of sememes we use is 1,335.

We use pretrained word embeddings of MWEs (needed for training in the MWE similarity task) and constituents, which are trained using GloVe \citep{pennington2014glove} on the Sogou-T corpus\footnote{Sogou-T is a corpus of web pages containing 2.7 billion words. \url{https://www.sogou.com/labs/resource/t.php}}. 
We also utilize pretrained sememe embeddings obtained from the results of a sememe-based word representation learning model\footnote{\url{https://github.com/thunlp/SE-WRL-SAT}} \citep{niu2017improved}.

And we build a dataset consisting of 51,034 Chinese MWEs, each of which and its two constituents are annotated with sememes in HowNet and have pretrained word embeddings simultaneously. 
We randomly split the dataset into training, validation and test sets in the ratio of $8:1:1$.



\subsection{Experimental Settings}


\paragraph{Baseline Methods} 
We choose several typical SC models as the baseline methods, including: (1) ADD and MUL, the simple additive and element-wise multiplicative models \citep{Mitchell2008}; (2) RAE, the recursive autoencoder model \citep{Socher2011}; (3) RNTN, the recursive neural tensor network \citep{Socher2013}; (4) TIM, the tensor index model \citep{Zhao2015};
and (5) SCAS-S, the ablated version of our SCAS model which removes sememe knowledge\footnote{SCAS-S is very similar to RAE, and the only difference between them is that the former concatenates the embeddings of two constituents while the latter chooses addition.}. 
These baseline methods range from the simplest additive model to complicated tensor-based model, all of which take no knowledge into consideration.

\paragraph{Combination Rules}
For simplicity, we divide all the MWEs in our dataset into four combination types, i.e., adjective-noun (Adj-N), noun-noun (N-N), verb-noun (V-N) and other (Other), whose instance numbers are 1302, 8276, 4242 and 37214 respectively. 
And we use the suffix +R to signify integrating combination rules into the model.

\paragraph{Hyper-parameters and Training}
The dimension of word and sememe embeddings $d$ is empirically set to $200$. 
$h_r$ in Eq. \eqref{eq:rule} is simply set to 5 for all the four combination types. 
The regularization parameter $\lambda$ is $10^{-4}$ , and $k$ in Eq. \eqref{eq:loss2}
is 100. 
As for training, we use Stochastic Gradient Descent (SGD) for optimization. The learning rate is initialized to 0.01 and 0.2 for the two tasks respectively, and decays by 1$\%$ every iteration. 
During training, word embeddings of an MWE's constituents are frozen while the sememe embeddings are fine-tuned.
For the baseline methods, they all use the same pre-trained word embeddings as our model and their hyper-parameters are tuned to the best on the validation set. 
We also use SGD to train them. 




\subsection{MWE Similarity Computation}
In this subsection, we evaluate our sememe-incorporated SC models and baseline methods on an intrinsic task, MWE similarity computation.   
\subsubsection*{Evaluation Datasets and Protocol}
We use two popular Chinese word similarity datasets, namely WordSim-240 (WS240) and WordSim-297 (WS297) \citep{Chen2015cwe}, and a newly built one, COS960 \citep{huang2019COS960},
all of which consist of word pairs together with human-assigned similarity scores. 
The first two datasets have 86 and 97 word pairs appearing in our MWE dataset respectively, and their human-assigned similarity scores are based on relatedness.
On the other hand, COS960 has 960 word pairs and all of them are in our MWE dataset. Moreover, its similarity scores are based on similarity.
We calculate the Spearman’s rank correlation coefficient between cosine similarities of word pairs computed by word embeddings of SC models and human-annotated scores. 





\subsubsection*{Experimental Results}
\begin{table}[htbp]
\resizebox{\linewidth}{!}{
  \centering
    \begin{tabular}{ccccc}
    \toprule
    Framework & Method & WS240& WS297 & COS960 \\
    \midrule
    \multirow{5}[2]{*}{$f(\textbf{w}_1,\textbf{w}_2)$} & ADD   & 50.8 & 53.1  & 49.1 \\
          & MUL   & 19.6  & 21.6  & $-$3.9 \\
          & TIM   & 47.4  & 54.2  & 50.5 \\
          & RNTN  & 42.5  & 53.6  & 55.8 \\
          & RAE   & 61.3  & 59.9  & 59.6 \\
          & SCAS-S   & 61.4  & 57.0  & 60.1 \\
    \midrule
    \multirow{2}[2]{*}{$f(\textbf{w}_1,\textbf{w}_2,K)$} & SCAS  & 60.2 & 60.5 & 61.4 \\
          & SCMSA  & \textbf{61.9}  & 58.7  & 60.5 \\
    \midrule
    \multirow{2}[2]{*}{$f(\textbf{w}_1,\textbf{w}_2,K,R)$} & SCAS+R & 59.0 &  60.8 &   \textbf{61.8}\\
          & SCMSA+R &  61.4 & \textbf{61.2} & 60.4 \\
    \bottomrule
    \end{tabular}%
}
  \caption{Spearman's rank correlation coefficient ($\rho \times 100$) between similarity scores assigned by compositional models with human ratings.}
  \label{tab:spearman}
\end{table}%

The experimental results of MWE similarity computation\footnote{Before training, we remove the MWEs which are in these three datasets from the training set.} are listed in Table \ref{tab:spearman}. We can find that:

(1) By incorporating sememe knowledge, our two SC models SCAS and SCMSA both achieve overall performance enhancement, especially on the COS960 dataset which has the largest size and reflects true word similarity. This result can prove the effectiveness of sememe knowledge in modeling SC. 
Although SCAS-S even performs better than SCAS on WS240, which is presumably because too few word pairs are used, SCAS significantly outperforms SCAS-S on the other two datasets.

(2) After further integrating combination rules, our two SC models basically produce better performance except on WS240, which can demonstrate the usefulness of combination rules to some extent.

(3) By comparing our two models SCAS and SCMSA, as well as their variants SCAS+R and SCMSA+R, we find no apparent advantage of attention-considered SCMSA over simple SCAS. We attribute it to insufficient training because SCMSA has more parameters. 

(4) Among the baseline methods, MUL performs particularly poorly on all the three datasets. 
Although \citet{Mitchell2008} report that multiplicative model yields better results than additive model based on distributional semantic space (SDS) word embeddings, we find it cannot fit the word embeddings obtained by currently popular methods like GloVe, which is consistent with the findings of previous work \citep{Zhao2015}.




\subsection{MWE Sememe Prediction}
According to the conclusion of the confirmatory experiment in Sec. \ref{sec:sec2}, the sememes of a word (or an MWE) can finely depict the semantics of the word (MWE). 
On the other hand, the high-quality embedding of a word (MWE) is also supposed to accurately represent the meaning of the word (MWE).
Therefore, we believe that the better the embedding is, the better sememes it can predict.
More specifically, whether an SC model can predict correct sememes for MWEs reflects the SC model's ability to learn the representations of MWEs.
Correspondingly, we regard MWE sememe prediction as a credible extrinsic evaluation of SC models. 



\subsubsection*{Evaluation Dataset and Protocol}
We use the above-mentioned test set for evaluation. 
As for the evaluation protocol, we adopt mean average precision (MAP) and F1 score following previous sememe prediction works \citep{xie2017lexical,qi2018cross}. 
Since our SC models and baseline methods yield a score for each sememe in the whole sememe set, we pick the sememes with scores higher than $\delta$ to compute F1 score, where $\delta$ is a hyper-parameter and also tuned to the best on the validation set.


\subsubsection*{Overall Results}
\begin{table}[!thbp]
\footnotesize
\resizebox{\linewidth}{!}{
  \centering
    \begin{tabular}{cccc}
    \toprule
    \multicolumn{1}{c}{\multirow{2}[4]{*}{Framework}} & \multicolumn{1}{c}{\multirow{2}[4]{*}{Method}} 
    & \multicolumn{2}{c}{Sememe Prediction} \\
\cmidrule{3-4}          &       & MAP   & F1 Score \\
    \midrule
    \multirow{5}{*}{$f(\textbf{w}_1,\textbf{w}_2)$} & ADD   & 40.7  &  23.2 \\
          & MUL   & 11.2  & 0.3 \\
          & TIM   & 46.8  & 35.3  \\ 
          & RNTN  & 47.7  & 35.3 \\
          & RAE   & 44.0  & 30.8 \\
          & SCAS-S   & 39.0  & 27.9 \\
\midrule
    \multirow{2}{*}{$f(\textbf{w}_1,\textbf{w}_2,K)$} & SCAS  & 52.2  & 41.3 \\
          & SCMSA  & 55.1 &  43.4\\
    \midrule
    \multirow{2}{*}{$f(\textbf{w}_1,\textbf{w}_2,K,R)$} & SCAS+R & 56.8  & \textbf{ 46.1} \\
          & SCMSA+R & \textbf{58.3} &  46.0\\
    \bottomrule
    \end{tabular}
}
\caption{Overall MWE sememe prediction results of all the models.}
\label{tab:sememe_prediction}
\end{table}

The overall sememe prediction results are exhibited in Table \ref{tab:sememe_prediction}. We can observe that:

(1) The effectiveness of sememe knowledge in modeling SC is definitively proved again by comparing our sememe-incorporated SC models with baseline methods, especially by the comparison of SCAS and its sememe-ablated version SCAS-S. 
Besides, the combination rule-integrated variants of our models perform better than corresponding original models, 
which makes the role of combination rules recognized more obviously.

(2) Our two models considering mutual attention, namely SCMSA and SCMSA+R models, produce considerable improvement by comparison with SCAS and SCAS+R models, which manifests the benefit of mutual attention mechanism. 

(3) MUL still performs the worst, which is consistent with the results of the last experiment. 




\subsubsection*{Effect of SCD}
In this experiment, we explore the effect of SCD (in Sec. \ref{sec:sec2}) on sememe prediction performance. 
We split the test set into four subsets according to MWE's SCD, which is computed by the sememe-based SCD methods in Table \ref{tab:compositionlity_score}. 
Then we evaluate sememe prediction performance of our models on the four subsets. 
From the results shown in Table \ref{tab:sp_comp_score}, we find that:

\begin{table}[!thbp]
\small
  \centering
    \begin{tabular}{lcccc}
    \toprule
    \multicolumn{1}{c}{\multirow{2}[4]{*}{Method}} & \multicolumn{4}{c}{SCD} \\
\cmidrule{2-5}          & 3     & 2     & 1     & 0 \\
    \midrule
    SCAS  & 88.4  & 63.8  & 46.9  & 13.3 \\
    SCAS+R & \textbf{95.9}  & 69.8  & 50.6  & 14.3  \\
    SCMSA  & 85.3  & 66.1  & 51.5  & \textbf{16.1} \\
    SCMSA+R & 91.2  & \textbf{71.2}  & \textbf{53.3}  & 14.5  \\
    \bottomrule
    \end{tabular}
    \caption{Sememe prediction MAP of our models on MWEs with different SCDs. The numbers of MWEs with the four SCDs are 180, 2540, 1686 and 698 respectively. }
  \label{tab:sp_comp_score}%
\end{table}

(1) MWEs with higher SCDs have better sememe prediction performance, which is easy to explain. MWEs with higher SCDs possess more meanings from their constituents, and consequently, SC models can better capture the meanings of these MWEs. 

(2) No matter integrating combination rules or not, our mutual attention models perform better than the aggregated sememe models, other than on the subset of SCD 3. 
According to previous SCD formulae, an MWE whose SCD is 3 has totally the same sememes as its constituents. 
That means in sememe prediction, each sememe of its constituents is equally important and should be recommended to the MWE. 
SCAS model simply adds all the sememes of constituents, which fits the characteristics of MWEs whose SCDs are 3. 
Thus, SCAS model yields better performance on these MWEs.



\subsubsection*{Effect of Combination Rules}
In this experiment, we investigate the effect of combination rules on sememe prediction performance. Table \ref{tab:sp_combine_rule} shows the MAPs of our models on MWEs with different combination rules. 



\begin{table}[!thbp]
\small
  \centering
    \begin{tabular}{lcccc}
    \toprule
          & Adj-N & N-N   & V-N   & Other \\
    \midrule
    Average SCD & 1.52  & 1.65  & 1.37  & 1.38 \\
    \midrule
    SCAS & 61.4  & 64.9  & 55.5  & 48.2 \\
    SCAS+R & 63.1  & 68.7  & 61.0    & 53.0 \\
    SCMSA & 59.6  & 66.2  & 58.8  & 51.8 \\
    SCMSA+R & 62.1  & 69.4  & 60.7  & 55.0 \\
    \bottomrule
    \end{tabular}%
    \caption{Sememe prediction MAP of our models on MWEs with different combination rules and average SCDs of the four subsets. The numbers of MWEs with the four combination rules are 157, 893, 443 and 3,611 respectively.}
  \label{tab:sp_combine_rule}%
\end{table}%

\begin{table*}[t!]
\resizebox{\linewidth}{!}{
  \centering
    \begin{tabular}{llccccc}
    \toprule
    \multicolumn{2}{c}{Words} & \multicolumn{5}{c}{Sememes} \\
    \midrule
    \multicolumn{2}{l}{\quad \textcolor{red}{参} (join; ginseng; impeach)} & \multicolumn{5}{c}{从事$|$engage,纳入$|$include,花草$|$FlowerGrass,药物$|$medicine,控告$|$accuse,警$|$police,政$|$politics} \\
    \midrule
    \multicolumn{2}{l}{\quad\textcolor{red}{参}\textcolor{blue}{战} (enter a war)} & \multicolumn{5}{c}{\textbf{争斗$|$fight}, \textbf{军$|$military}, \textbf{事情$|$fact}, \textbf{从事$|$engage}, 政$|$politics} \\

    \midrule
    \multicolumn{2}{l}{\textcolor{blue}{丹}\textcolor{red}{参} (red salvia)} & \multicolumn{5}{c}{\textbf{药物$|$medicine}, \textbf{花草$|$FlowerGrass}, 红$|$red, 生殖$|$reproduce, 中国$|$China} \\
    \bottomrule
    \end{tabular}%
}  
  \caption{An example of sememe prediction when two MWEs share the same constituent 参. Top5 predicted sememes are presented in the second and third lines. Bold sememes are correct.}
  \label{tab:case_study}%
\end{table*}%

We find that integrating combination rules into SC models is beneficial to sememe prediction of MWEs with whichever combination rule. 
In addition, sememe prediction performance varies with the combination rule. 
To explain this, we calculate the average SCDs of the four subsets with different combination rules, and find that their sememe prediction performance is positively correlated with their average SCDs basically (the average Pearson's correlation coefficient of different models is up to 0.87). This conforms to the conclusion of the last experiment.


\subsubsection*{Case Study}


Here, we give an example of sememe prediction for MWEs comprising polysemous constituents, to show that our model can capture the correct meanings of constituents in SC.

As shown in Table \ref{tab:case_study}, the Chinese word 参 has three senses including ``join'', ``ginseng'' and ``impeach'', and these meanings are represented by their different sememes.  
For the MWE 参战, whose meaning is ``enter a war'', 参 expresses its first sense ``join''. In the top 5 predicted sememes of our SC model, the first four are the sememes annotated in HowNet,  including the sememe ``从事$|$engage'' from 参. In addition, the fifth sememe ``politics'' is also related to the meaning of the MWE.
For another MWE 丹参, which means ``red salvia'', a kind of red Chinese herbal medicine resembling ginseng, the meaning of 参 here is ``ginseng''. Our model also correctly predicts the two sememes ``药物$|$medicine'' and ``花草$|$FlowerGrass'', which are both annotated to 参 in HowNet. In addition, other predicted sememes given by our model like ``红$|$red'' and ``中国$|$China'' are also reasonable.

This case demonstrates that our sememe-incorporated SC model can capture the correct meanings of an MWE's constituents, especially the polysemous constituents. And going further, sememe knowledge is beneficial to SC and our SC model can take advantage of sememes.





\section{Related Work}
\subsection{Semantic Compositionality}
Based on the development of distributional semantics, vector-based SC modeling has been extensively studied in recent years. 
Most existing work concentrates on using better compositionality functions. 
\citet{Mitchell2008} first make a detailed comparison of several simple compositionality functions including addition and element-wise multiplication. 
Then various complicated models are proposed in succession, such as vector-matrix models \citep{baroni2010nouns,Socher2012}, matrix-space models \citep{yessenalina2011compositional,grefenstette2011experimental} and tensor-based models \citep{grefenstette2013multi,VandeCruys2013,Socher2013}.

There are also some works trying to integrate combination rules into semantic composition models \citep{Blacoe2012,Zhao2015,Kober2016,Weir2016}. 
But few works explore the role of external knowledge in SC. 
\citet{Zhu2016} incorporate prior sentimental knowledge into LSTM models, aiming to improve sentiment analysis performance of sentences.
To the best our knowledge, there is no work trying to take account of general linguistic knowledge in SC, especially for the MWE representation learning task.

\subsection{Sememes and HowNet}
HowNet, as the most well-known sememe KB, has attracted wide research attention. Previous work applies the sememe knowledge of HowNet to various NLP applications, such as word similarity computation \citep{liu2002}, word sense disambiguation \citep{gan2000annotating,zhang2005chinese,duan2007word}, sentiment analysis \citep{zhu2006semantic,dang2010method,fu2013multi}, word representation learning \citep{niu2017improved}, language modeling \citep{gu2018language}, lexicon expansion \citep{zeng2018chinese} and semantic rationality evaluation \citep{liu2018evaluating}.

To tackle the challenge of high cost of annotating sememes for new words, \citet{xie2017lexical} propose the task of automatic sememe prediction to facilitate sememe annotation. And they also propose two simple but effective models. \citet{jin2018incorporating} further incorporate Chinese character information into their sememe prediction model and achieve performance boost. \citet{li2018sememe} explore the effectiveness of words' descriptive text in sememe prediction task. In addition, \citet{qi2018cross} make the first attempt to use cross-lingual sememe prediction to construct sememe KBs for other languages.

\section{Conclusion and Future Work}
In this paper, we focus on utilizing sememes to model semantic compositionality (SC).
We first design an SC degree (SCD) measurement experiment to preliminarily prove the usefulness of sememes in modeling SC. 
Then we make the first attempt to employ sememes in a typical SC task, namely MWE representation learning. 
In experiments, our proposed sememe-incorporated models achieve impressive performance gain on both intrinsic and extrinsic evaluations in comparison with baseline methods without considering external knowledge.

In the future, we will explore the following directions: 
(1) context information is also essential to MWE representation learning, and we will try to combine both internal information and external context information to learn better MWE representations; 
(2) many MWEs lack sememe annotation and we will seek to calculate an MWE's SCD when we only know the sememes of the MWE's constituents;
(3) our proposed models are also applicable to the MWEs with more than two constituents and we will extend our models to longer MWEs;
(4) sememe is universal linguistic knowledge and we will explore to generalize our methods to other languages.


\section*{Acknowledgments}
This research is jointly supported by the Natural Science Foundation of China (NSFC) project under the grant No. 61661146007 and the NExT++ project, the National Research Foundation, Prime Minister’s Office, Singapore under its IRC@Singapore Funding Initiative.
Moreover, it is also funded by the NSFC and the German Research Foundation (DFG) in Project Crossmodal Learning, NSFC 61621136008 / DFG TRR-169, as well as the NSFC project under the grant No. 61572273.
Furthermore, we thank the anonymous reviewers for their valuable comments and suggestions.

\bibliography{acl2019}
\bibliographystyle{acl_natbib}

\end{CJK}
\end{document}